# A white-box SVM framework and its swarm-based optimization for supervision of toothed milling cutter through characterization of spindle vibrations

Tejas Y. Deo[1], B. B. Deshmukh[2], Keshav H. Jatakar[3], Kamlesh M. Chhajed[4], S. S. Pardeshi[5], R. Jegadeeshwaran[6], Apoorva N. Khairnar[5], Hrushikesh S. Khade[5], A. D. Patange[5]

[1]P.E.S. Modern College of Engineering, Pune, Maharashtra 411005, India

[2]Walchand Institute of Technology, Solapur, Maharashtra 413006, India

[3]Rizvi College of Engineering, Bandra (W), Mumbai, Maharashtra 400050, India

[4]SVKM NMIMS Global University, Dhule, Maharashtra 424 001, India

[5]College of Engineering Pune, Pune, Maharashtra 411005, India

[6]Vellore Institute of Technology, Chennai, Tamil Nadu 600127, India

Corresponding author: Abhishek D. Patange, Email: abhipatange93@gmail.com

**Abstract:**

In this paper, a white-Box support vector machine (SVM) framework and its swarm-based optimization is presented for supervision of toothed milling cutter through characterization of real-time spindle vibrations. The anomalous moments of vibration evolved due to in-process tool failures (i.e., flank and nose wear, crater and notch wear, edge fracture) have been investigated through time-domain response of acceleration and statistical features. The Recursive Feature Elimination with Cross-Validation (RFECV) with decision trees as the estimator has been implemented for feature selection. Further, the competence of standard SVM has been examined for tool health monitoring followed by its optimization through application of swarm based algorithms. The comparative analysis of performance of five meta-heuristic algorithms (Elephant Herding Optimization, Monarch Butterfly Optimization, Harris Hawks Optimization, Slime Mould Algorithm, and Moth Search Algorithm) has been carried out. The white-box approach has been presented considering global and local representation that provides insight into the performance of machine learning models in tool condition monitoring.

## 1. Introduction

Machining processes involve creating a workpiece with the desired shape. This is achieved by removing metal from the piece using tools in different ways. This includes the raw material i.e., the stock from which the workpiece is cut out, a cutting tool with the required properties, and the machine to carry out the machining process [1]. Machining processes are classified into three principal processes viz. turning, drilling, milling. The machining tools differ with these processes. Turning includes material removal outside the diameter on a lathe machine. Material removal is achieved in a precise way with a stationary cutting tool applied to a workpiece fixed onto the chuck, which can revolve at set speeds [2]. Drilling involves creation of holes in the workpiece using a drill bit. The tool enters the workpiece and cuts out hole like shape equivalent to the diameter of the tool. Milling is another key operation in machining [3]. An axially asymmetric workpiece is attached to a fixture and then mounted on the milling machine, either vertical or horizontal, as required. The cutter, attached to the machine, rotates and removes material as the workpiece is passed through. All machines have provision for cooling liquid to keep temperatures low during the machining [4]. Due to friction between the workpiece and the cutting tool, heat is generated and needs to be controlled to avoid reaching critical temperatures and softening and accelerated wear of tools. The pressure applied due to the stationary tool and rotating workpiece or vice versa shears the metal from the piece. Cut-out material is in form of chips and the mechanics of chip formation play an important role in accuracy & precision in the process and in increasing the life of the tool [5].

Milling is a versatile machining process. We have used a vertical milling cutter in this research. It has multiple degrees of freedom, which is not found in drilling and turning. Generally used for flat surfaces, milling can also be used for irregular surfaces such as inclined and curved surfaces [6]. As mentioned before, milling includes a rotating multi-toothed cutter that is fed into the moving workpiece. The workpiece is mounted on a bed, which can be fed in X, Y, and Z axes manually. This can also be done with a power feed to deliver a smoother finish. The head of the milling cutter may also be tilted from front to back and side to side. Although this comes with the possibility of errors in the piece, it can be avoided by adjusting the drift in the tilt regularly. The milling machine uses a multi-point tool, also known as a tipped tool [7]. The multiple cutting edges serve advantages like better surface finish, faster material removal, and lower cutting temperatures. The nature of cutting is intermittent, rather than continuous as in turning. Intermittent cutting wears out the tool faster but gives a higher speed of machining. Recently M. Kuntoğlu et al. presented a comprehensive review detailing sensors and signal processing schemes in machining activities [8].

Advancements in technology have simultaneously increased the demand for precision in products and the methods to produce the same [9]. The increasing need for precision and product quality has boosted the development of tool condition monitoring (TCM) along with automatic manufacturing processes [10]. The industry is looking to minimize the reject rate of workpiece if not avoid it completely. Machine Learning (ML) and Artificial Intelligence (AI) are seeing huge contributions in TCM to avoid tool failures. ML, the subset of AI, involves analysing data to recognize patterns that lead to outcomes and automates analytical model building [11-14]. With classification and regression being the main branches in ML, this research only focuses on classification as the aim is to supervise the condition of a tool. Single class, multi-class classification algorithms in ML are commonly used today, both in everyday applications like spam mails and in research to classify the fatality of cancerous patients. Support Vector Machine algorithms originated from a geometric idea about the classification problem. The training data points are mapped in space to maximize the gap between the two classes in binary classification. Testing data points are mapped in this model space and classified according to the side of the gap that they fall in. Using kernel trick, SVMs can also be used for multiclass classification with a non-linear boundary where the inputs are mapped onto high-dimensional feature spaces. SVMs are also used with unlabelled data with the support vector clustering algorithm, however, in this study, we have labelled data. Optimizing the parameters of a real-world problem is a complex task. Meta-heuristics may provide a sufficiently good solution to an optimization problem but do not guarantee to give global optimum solutions. They often find good solutions with less computational effort than other optimization algorithms, iterative methods, or simple heuristics. In this paper, traditional optimization techniques like Sequential Minimal Optimization (SMO), Grid Search CV, and Random Search CV are compared with different meta-heuristic optimizers. There are a large number of studies in the field of TCM using machine learning and artificial intelligence. However, the focus of these papers emphasizes to implement models or optimizations to achieve better classification accuracy and is mostly black-box models. We intend to explain the classifications, both correct and incorrect. Researchers would be interested in knowing the features that can avoid a misclassification instead of those that contribute to one, as contrasting explanations are important. The lost predictive performance with interpretable models can be avoided with model-agnostic methods.

## 2. **Literature Review**

The profit-based nature of manufacturing industry calls for highly efficient processes along with product quality maintenance. For this purpose, higher cutting speeds are used by manufacturers quite frequently. However, this gives rise to stress concentration and temperature rise in the tool regions which adversely affects product quality and tool health. Such effects on the health of a tool result in a high demand for tool condition monitoring (TCM). In the past two decades, TCM has gained a lot of attention in industrial as well as academic research. The recent developments in machine learning (ML) and easily accessible open-source tools like codes and functions have resulted in a huge increase in the application of ML in TCM. This field demands predictive TCM using ML algorithms. Selection of classification technique depends on a number of factors like type and size of data, computational effort, number of features, kind of output, etc. Artificial Neural Networks (ANN), Kernel-based Support Vector Machine (SVM), Decision Trees (DT) and Random Forest and K-nearest Neighbours (KNN) are a few commonly used techniques in ML-based TCM. The study carried out by Mohanraj et al. [15] designed SVM, Multi-layered Perceptron (MLP), Kernel-based Bayes & DT followed by comparative study based on confusion matrix/classification accuracy. Further classification of flank wear in different phases has been demonstrated using designed algorithms for single-point cutting tool. Based on the comparison parameters, SVM and DT were found to perform quite well in forecasting the fault, however, overfitting was observed when DTs were used. Hence, SVM is considered the best classifier in this study. SVMs can be tuned according to the problem statement and can be used on small datasets as well. The current study inspires from the comparative study carried out by Mohanraj et al. [15] and thus advocates explainable (white-box) design of SVM classifiers and its optimization. However, the results of SVM depend largely upon the hyper-parameters. Hence, optimization of these parameters is required which is usually done by evaluating a performance parameter like error, accuracy, etc. The SVM parameters – regularization parameter $C$, kernel parameter $\gamma$ and kernel function – are usually optimized using GridSearchCV technique. However, this technique is not suitable for high dimensional data and requires a lot of computational effort. Another optimization technique – Sequential Minimal Optimization (SMO) has exhibited good convergence properties. It iteratively selects subsets of size two and optimises the target function. Recently, meta-heuristic algorithms have gained attention in optimization problems. Unlike other optimization techniques, meta-heuristic algorithms attempt to mimic the naturally occurring randomness. They can be categorised as nature-inspired and non-nature-inspired, where nature-inspired algorithms are further classified into Evolutionary, Physics-based, Swarm-based, Human-based and Chemical-based. The table 1 summaries research investigations in which contemporary swarm-based optimizers are employed for solving typical problems from diverse domains.

**Table 1:** Summary of applications of contemporary swarm-based optimizers

| Sr. No. | Reference | Title | Application |
|---|---|---|---|
| | ***Elephant Herding Optimizer*** | | |
| 1 | Li, W. et al. (2020) [16] | Learning-based elephant herding optimization algorithm for solving numerical optimization problems | Solving Optimization problems |
| 2 | Nayak, M. et al. (2020) [17] | Elephant herding optimization technique based neural network for cancer prediction | Optimization of Neural Network for Cancer Prediction |
| 3 | Li, J. et al. (2020) [18] | Real-Time Predictive Control for Chemical Distribution in Sewer Networks Using Improved Elephant Herding Optimization | Chemical Distribution in Sewer Networks |
| 4 | Dhillon, S. S. et al. (2020) [19] | Automatic Generation Control of Interconnected Power Systems Using Elephant Herding Optimization | Power Systems |
| 5 | Li, W., & Wang, G. G. (2021) [20] | Improved elephant herding optimization using opposition-based learning and K-means clustering to solve numerical optimization problems | Numerical optimization problems using K-means clustering |
| | ***Monarch Butterfly Optimizer*** | | |
| 6 | Yi, J. H. et al. (2019) [21] | Using Monarch Butterfly Optimization to Solve the Emergency Vehicle Routing problem with Relief materials in Sudden Disasters | To solve Vehicle Routing Problem |
| 7 | Bacanin, N. et al. (2020) [22] | Monarch Butterfly Optimization Based Convolutional Neural Network Design | Convolutional neural network |
| 8 | Singh, P. et al. (2020) [23] | Multi-criteria decision-making monarch butterfly optimization for optimal distributed energy resources mix in distribution networks | Energy resources mix in distribution networks |
| 9 | Ghanem, W. A., & Jantan, A. (2020) [24] | Training a Neural Network for Cyberattack Classification Applications Using Hybridization of an Artificial Bee Colony and Monarch Butterfly Optimization | Cyberattack classification |
| 10 | Dorgham, O. M. et al. (2021) [25] | Monarch butterfly optimization algorithm for computed tomography image segmentation | Image segmentation |
| | ***Harris Hawks Optimizer*** | | |
| 11 | Houssein, E. H. et al. (2020) [26] | A novel hybrid Harris hawks optimization and support vector machines for drug design and discovery | Drug design and discovery |
| 12 | Elgamal, Z. M. et al. (2020) [27] | An Improved Harris Hawks Optimization Algorithm With Simulated Annealing for Feature Selection in the Medical Field | Feature selection in Medical field |
| 13 | Golilarz, N. A. et al. (2019) [28] | A New Automatic Method for Control Chart Patterns Recognition Based on ConvNet and Harris Hawks Meta Heuristic Optimization Algorithm | Pattern recognition using ConvNet |
| 14 | Bao, X. et al. (2019) [29] | A Novel Hybrid Harris Hawks Optimization for Color Image Multilevel Thresholding Segmentation | Image segmentation |
| 15 | Tikhamarine, Y. et al. (2020) [30] | Rainfall-runoff modelling using improved machine learning methods: Harris hawks optimizer vs. particle swarm optimization | Rainfall-runoff modelling |
| | ***Slime Mould Algorithm*** | | |
| 16 | Mostafa, M. et al. (2020) [31] | A new strategy based on slime mould algorithm to extract the optimal model | Solar PV Panel |

| | | parameters of solar PV panel | |
|---|---|---|---|
| 17 | Zubaidi, S. L. et al. (2020) [32] | Hybridised Artificial Neural Network Model with Slime Mould Algorithm: A Novel Methodology for Prediction of Urban Stochastic Water Demand | Prediction of Water Demand |
| 18 | Kumar, C. et al. (2020) [33] | A new stochastic slime mould optimization algorithm for the estimation of solar photovoltaic cell parameters | To estimate solar photovoltaic cell parameters |
| 19 | Liu, L. et al. (2021) [34] | Performance optimization of differential evolution with slime mould algorithm for multilevel breast cancer image segmentation | Breast cancer image segmentation |
| 20 | Tiachacht, S. et al. (2021) [35] | Inverse problem for dynamic structural health monitoring based on slime mould algorithm | Structural health monitoring |
| | ***Moth Search Algorithm*** | | |
| 21 | Razmjooy, N. et al. (2021) [36] | A New Design for Robust Control of Power System Stabilizer Based on Moth Search Algorithm | Power system stabilizer |
| 22 | Shankar, K. et al. (2020) [37] | Deep neural network with moth search optimization algorithm based detection and classification of diabetic retinopathy images | Classification of Diabetic retinopathy images |
| 23 | Singh, P. et al. (2019) [38] | Moth Search Optimization for Optimal DERs Integration in Conjunction to OLTC Tap Operations in Distribution Systems | Distributed Systems |
| 24 | Carrasco, O. et al. (2019) [39] | Optimization of Bridges Reinforcements with Tied-Arch Using Moth Search Algorithm | Bridge Reinforcements |
| 25 | Han, X. et al. (2020) [40] | Efficient hybrid algorithm based on moth search and fireworks algorithm for solving numerical and constrained engineering optimization problems | Engineering optimization problems |

The iterative nature of biological evolution process is used for optimization in Evolutionary algorithms. Genetic Algorithm (GA), a type of Evolutionary algorithm, was the first proposed meta-heuristic algorithm. In this research SVMs were optimised using GA (GA-SVM) to predict bankruptcy. The study made use of GA-SVM in predicting the activity of BK-channels i.e., classification of medicinal chemistry and heart diseases in. Inspired by the collective social behaviour of swarms or communities such as schools of fish, herds of animals, flocks of birds, and colonies of insects are swarm intelligence algorithms. To name a few, Particle Swarm Optimization (PSO), Dragonfly algorithm (DA), Grey Wolf Optimization Algorithm (WOA) is commonly used swarm-based meta-heuristic algorithms. In addition to that, recently, Elephant Herding Optimization (EHO), Monarch Butterfly Optimization (MBO), Moth Search Algorithm, Harris Hawks Optimization (HHO) and Slime Mould Algorithm (SMA) have also been designed. An EHO-SVM approach for automatic electrocardiogram (ECG) signal classification has also been proposed. In this study, the features and parameters were optimized using EHO-SVM for accurate classification. Similarly, HHO-SVM integration has been utilised in extraction of best features for the classification of chemical descriptor selection and chemical compound activities in drug design and delivery. Furthermore, Mathematics-based optimization techniques include Sine Cosine Algorithm (SCA) and Singular Spectrum Analysis (SSA) as some of the remarkable algorithms. The application of these algorithms in rainfall and runoff forecast has been proposed. However, as proved by the No Free Lunch theorem, for a given problem statement, a single meta-heuristic algorithm cannot yield the best optimization results. Hence, the application of different meta-heuristic optimization techniques improves the scope of a research and makes the algorithms more suitable in specific areas.

## 3. Methodology and Contributions

The characterization of machining vibration pertaining tool faults has been carried out systematically. The features such as mean, median, shape factor, k-factor, mode, kurtosis, skewness, standard deviation, maximum, minimum and variance were computed. For information gain, feature-importance values extracted from a trained Decision Tree classifier justifies that some features are of greater importance than others. To select the best combination of features, a model-based selection technique called as Recursive Feature Elimination with Cross Validation has been employed. Removing features with high positive correlation and standardizing the data is necessary to avoid over-fitting. This final pre-processed data is utilized to test different kernels used in Support Vector Machines (SVM).

The parameters of Support Vector Classifier (SVC) with the kernel best suitable for the task, undergoes optimization via traditional methods like Grid Search CV, Random Search CV, and Sequential Minimal Optimization (SMO). Different Meta-heuristic optimizers (Swarm Intelligence based) have been used to further increase the 10-fold CV accuracy and to better optimize the SVM parameters. The results of all optimization algorithms have been inferred and compared. Being inherently interpretable and visually representable, Decision Tree is used to extract rules from underlying Machine Learning models using the Eli5 library. The highest and lowest performing models for both, optimized and vanilla SVMs, were selected for white-box approach. The contributions of current study are as follows:

- A ML-based classification approach for health monitoring of tipped milling cutter with the real-time signal acquisition has been demonstrated.
- Anomalous moments of vibration evolved due to tool failures (i.e., flank and nose wear, crater and notch wear, edge fracture, etc.) have been investigated.
- Recursive Feature Elimination with Cross Validation (RFECV) with Decision Trees as the estimator has been implemented for feature selection.
- The application of Swarm based algorithms for optimization of SVM classifier and hyper-parameter robustness analysis for health monitoring has been presented.
- Comparative analysis of performance of five meta-heuristic optimization algorithms alongside traditional optimization techniques has been carried out.
- The white-box approach has been presented considering global and local representation that provides insight into the performance of machine learning models in tool condition monitoring.

## 4. Experimentation and Data Collection

In this section, the discussion about the experimental setup, how the faults were characterized, and the data collection procedure has been presented.

### 4.1 Development of Setup

The setup used is defined by the following parameters. The accelerometer was mounted vertically on the spindle frame to capture any variation in spindle motion, especially those along the Z axis. Standard input parameters are used for machining. (Machining speed = 150m/min, Table feed = 50 m/min, Cutting depth = 0.35 mm). Figure 1 shows the experimentation setup & location of sensor.

- MTAB Compact Mill: CNC milling trainer
- MITSUBISHI M70: Controller, drives the Mill
- Workpiece: cast iron, C-shaped hollow cuboid, size 350x125x50 $mm^3$
- Milling tool: Dia. 40 mm, mounted inside the spindle
- PCB, Model: 352C03 ICP, piezoelectric type accelerometer, sensitivity factor 10 mV/gravitational acceleration, range ± 500 times gravitational acceleration
- NI Inc. National Instruments, Model: 9234 with four channels, Data acquisition card and connected to LabVIEW GUI

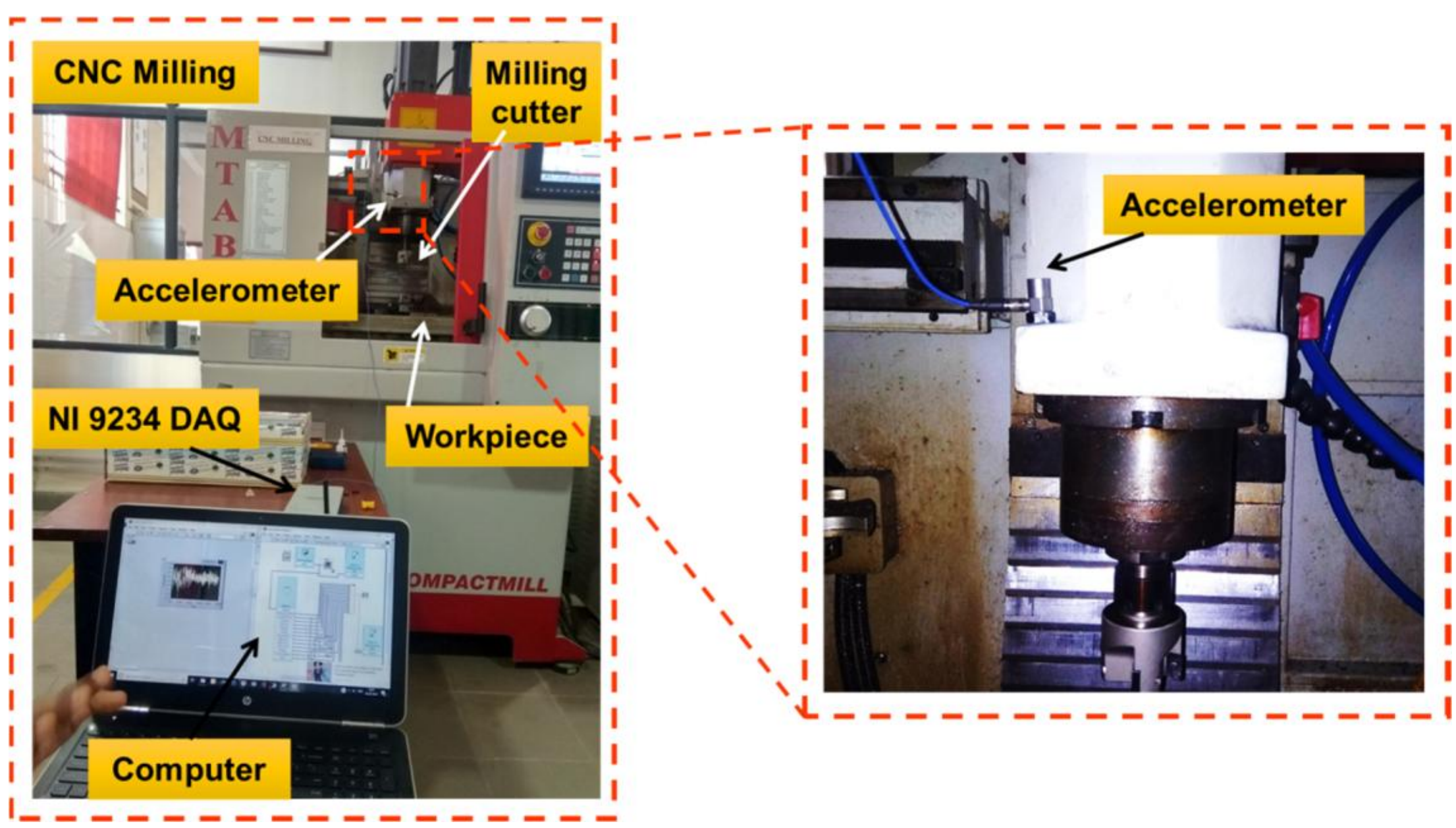

**Figure 1:** Experimentation setup & location of sensor

## 4.2 Characterization of Faults

A four carbide-coated insert milling cutter, with five faulty and one healthy configuration, is used. The faults induced were nose wear, flank wear, notch wear, crater wear, and edge fracture. In the first configuration, all inserts were healthy (new). For the next run, one insert was replaced with a faulty insert, and so on. The magnitude is the severity of the defective conditions with which the vibrations are read by the accelerometer. This magnitude is measured with the Conation Technologies, Model: SuXma-MOTO, an automatic metallurgical microscope, with a magnification range of 50X to 1000X.

## 4.3 Procedure for Data Collection

The data acquisition system (DAQ) is turned on and the following parameters are set using LabVIEW GUI. This procedure collects 50*12000 = 600,000 data points which can be used to build a sturdy model. The noise is filtered by rejecting the out-of-band signals using the NI-9234 DAQ, with both digital and analog filter combinations. There are a total of 6 classes for 6 different operations performed. The 6 different conditions used are Normal condition, Wear at nose radius, Notch wear, Wear at flank face, Crater wear, Edge fracture.

- Sampling frequency: 20 kHz
- Sample length: 12000 data points
- Number of samples per configuration: 50 (arbitrarily selected)

### 4.4 Data Representation

The above-collected data is represented in the time domain. The change in response as read by the accelerometer has been noted. For first configuration, Low amplitude and uniformity are observed which depicts the tool with all four inserts healthy. However, the cyclic and periodic variations in data can be seen in for all other configurations which correspond with the defective tool inserts. This gives insight into the dissimilarity between the tool configurations owing to analogs. In the machine learning, SVM classification, optimization and white-box approach is explained.

## 5. Machine Learning, SVM Classification, Optimization and White-box Approach

This section describes feature selection first, then classification through SVM and optimization of SVM using meta-heuristic algorithms and followed by Eli5: a way to facilitate model-agnostic approach. The flowchart of the current investigation is shown in Figure 2.

### 5.1 Data pre-processing

The t-Distributed Stochastic Neighbour embedding (t-SNE) is an unsupervised, non-linear technique primarily used for data exploration and visualizing high-dimensional data. It captures the intuition of the arrangement of data in a high-dimensional space. The risk of over-fitting increases with an increasing number of features. Feature importance is necessary to avoid longer training times for models and over-fitting. In Decision trees, the best performing features when close to the root to assess their significance. Model-based feature importance can improve performance on a given dataset. For example, in a decision tree at a node, the feature split that maximizes the information gain is the deciding feature for that node.

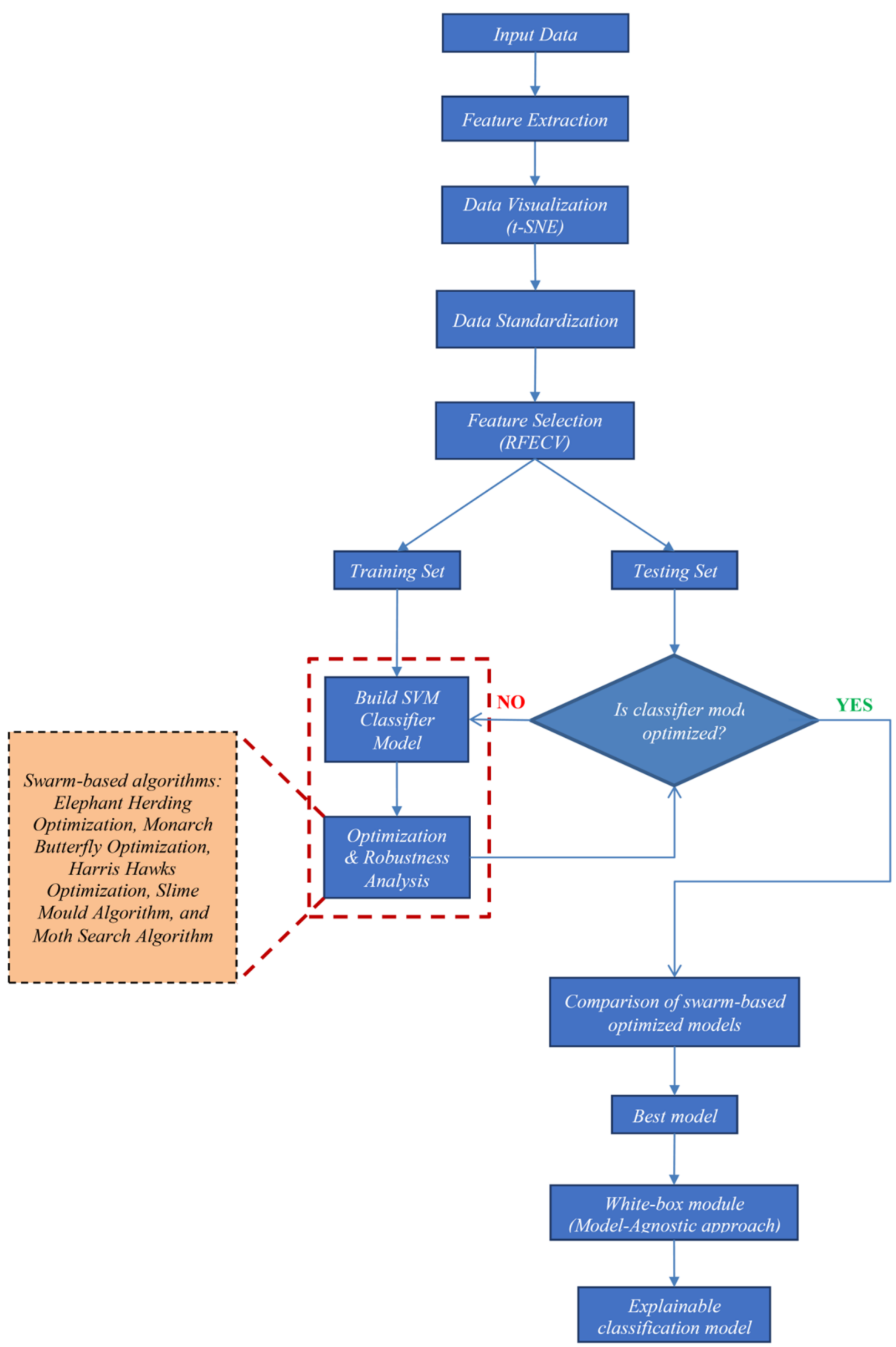

**Figure 2:** Flowchart of the current investigation

Decision Trees are stochastic in nature as the features are always randomly permuted at each split, even if the splitter is set to 'best'. Recursive feature elimination with cross-validation (RFECV) is a popular method to select features. RFE is a backward selection of predictors. With the entire set of predictors, we build a model and eliminate the least significant predictors. The model is built again with the new set of predictors, and we repeat this process. The user decides the size and number of predictor subsets which makes it a tuning parameter for RFECV. From the best-selected features (i.e. predictors), predictors having high positive correlation are removed. The final model is trained with the optimal subset size. The diverse nature of features calls for data standardization.

### 5.2 Classification through SVM and Optimization using Meta-heuristic Algorithms

This paper demonstrates the use of Support Vector Machines with Radial Basis Function (RBF) Kernel for multi-class classification problems. Support Vector Machines are a set of supervised learning methods used for classification, regression, and outlier detection. SVM's are chosen due to their following advantages:

- Its usefulness in high-dimensional spaces.
- Gives good results even when the number of dimensions is greater than the number of samples.
- Has a low space and time complexity as it uses a subset of training points in the decision function (called the support vectors)
- The decision function consists of different kernel functions that transform the non-separable data to separable data by converting the data points from m-dimension and X space to d-dimension and Z space.

Meta-heuristic optimizing algorithm use different search techniques to converge to the most optimum solution in the given range. These have gained popularity due to the interface of algorithms including simplicity, flexibility, and derivation free mechanism and avoiding local optima. Swarm based intelligence (SI) algorithms are one of the four categories of meta-heuristic optimization techniques with Physical based, Evolutionary based, and Human based being the other three. Swarm Intelligence, as the name suggests, gathers inspiration from the social intelligence of creatures. The search agents navigate in above methods which help in preservation of search information within the domain during integration [41]. The best solution so far is saved in the algorithm memory which can be used to compare the current best [42].

### 5.3 Eli5: A way to facilitate Model-Agnostic Approach

Model-Agnostic approach is a method that can be used to any machine learning model, from support vector machines to neural networks. This paper demonstrates the use of model agnostic method for model interpretation and extracting the rules that govern the model's decision. Eli5 is a python package that helps to debug machine learning classifiers and explain the reason behind their classification. There are two main ways of looking at a classification or regression problem from an interpretive point of view:

- Global Interpretation: To figure out how the model works globally and inspect the model parameters.
- Local interpretation: Inspect individual prediction of the model, to figure out the reason behind classification of a single datapoint.

Global inspection of the model is done using the "show_weights" function. This function determines the impact that a feature has towards model prediction. Another way to understand the global model is by using the Permutation models. The output (weight values) that the permutation model gives is similar to the feature importance values. It also shows the relative importance among the features. A feature is important if shuffling its values increases the model error. The local interpretation is done by using the "show_prediction" function. This paper demonstrates the use of decision tree classifiers to extract rules from the trained SVM models. The reason behind choosing decision tree classifiers is:

- Decision trees are inherently interpretable and give a visual representation.
- Eli5 is not applicable to SVMs with RBF kernels.
- The main job of the decision tree is to extract the rules followed by the SVM models to make a decision.
- The model agnostic process goes as follow:
- The vanilla and best optimized SVM models are used to make a prediction on the training and testing data.
- This predicted data acts as the input for the decision tree classifiers.
- Higher accuracy given by the decision tree on this predicted data underscores the fact that, they are able to demonstrate or interpret the working of SVM models on the original input data.

## 6. Results and Discussion

The results are presented in this section by supporting pertinent discussions starting with feature selection, then classification through SVM and optimization of SVM using meta-heuristic algorithms and followed White-box results considering global representation (Permutation Model) for best optimized SVM, global representation (Permutation Model) for vanilla SVM, local Representation for best optimized SVM, and local representation for vanilla SVM.

### 6.1 Feature Selection

The t-SNE plot of the original data i.e., raw data is shown in Figure 3. The plot suggests a lack of distinctiveness in the original data. Labels 0, 1, 2, 3, 4, 5 indicate Conditions A (Normal Condition), B (Wear at Nose Radius), C (Notch Wear), D (Wear at Flank Face), E (Crater Wear), and F (Edge Fracture) respectively. There are no missing values in the original dataset and so data imputation is not required.

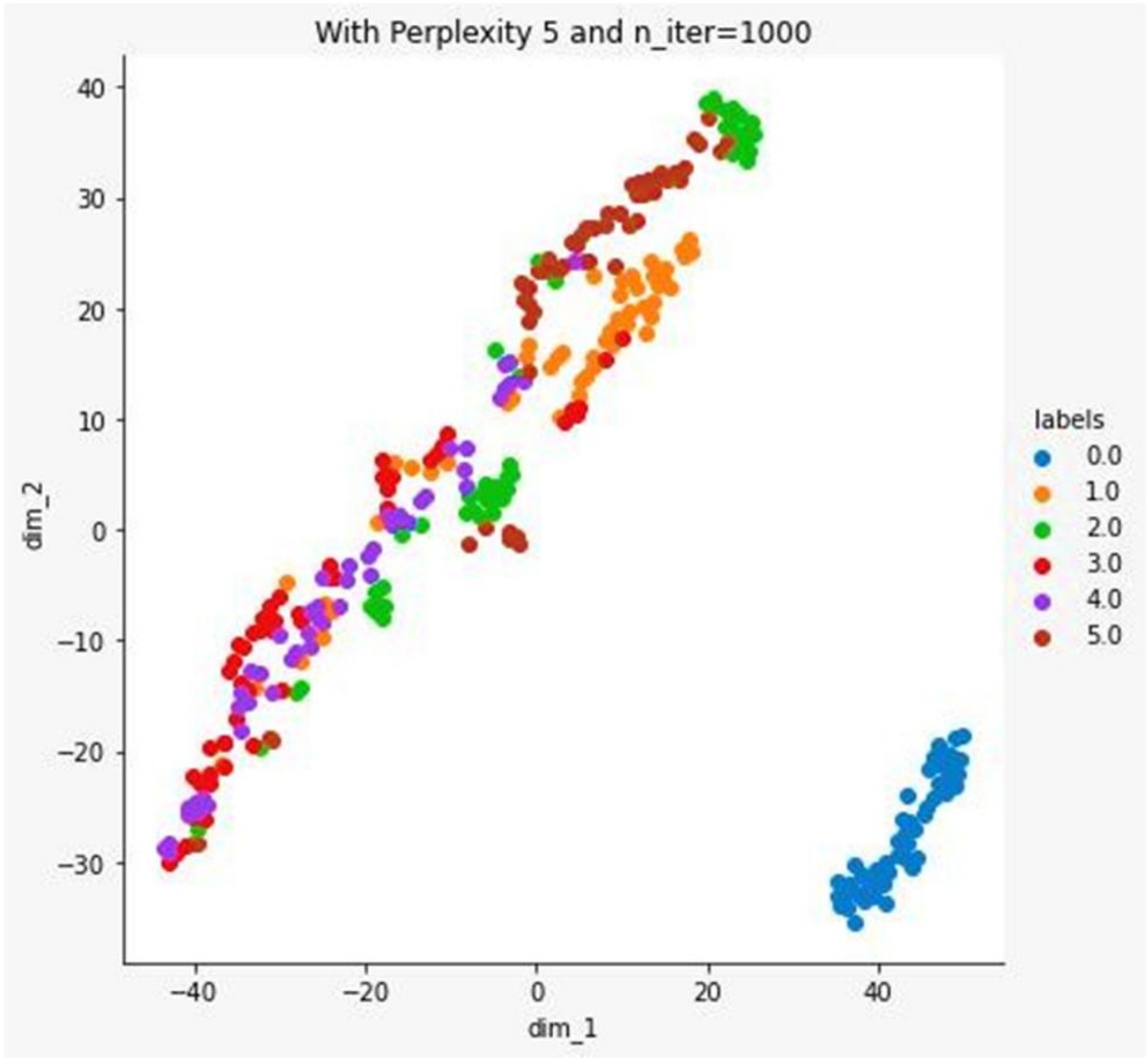

**Figure 3:** t-SNE plot of original data

Since the dataset consists of just 17 features, it is possible to train a Decision Tree Classifier to know the usefulness of the data. The "max_depth" argument of the Decision

Tree is 8 which are sufficient for overfitting given the size of the data. The classifier gives an accuracy of about 85% on average (as Decision Tree is a stochastic classifier because “random_state” is set to none). Extracting the feature importance values shows that some features are more important than others. Recursive Feature Elimination is a useful technique to select the top 13 features (Mean, Standard deviation, Variance, Kurtosis, Sum, Skewness, Max, Min, Range, RMS, Shape factor, k factor, Std error). The estimator used here is a Decision Tree Classifier. Plotting a correlation matrix is useful to remove features that have a high positive correlation (as shown in Figure 4). Having a high positive correlation between the features is harmful to the model as the presence of an anomaly in one feature will affect the rest. Shape factor, k factor, and Standard error are having a high positive correlation, and keeping any 1 feature (Shape factor) will suffice. Since Variance is the Standard Deviation square, squaring an anomaly if present is harmful during model training. Also, these features being highly correlated with each other, Variance is dropped. The final correlation matrix can be seen in Figure 5. Standardizing the data is necessary as some features have a higher range, which could get more importance during model training. The final 10 features selected for training the models are (Mean, Standard deviation, Kurtosis, Sum, Skewness, Max, Min, Range, RMS, Shape factor).

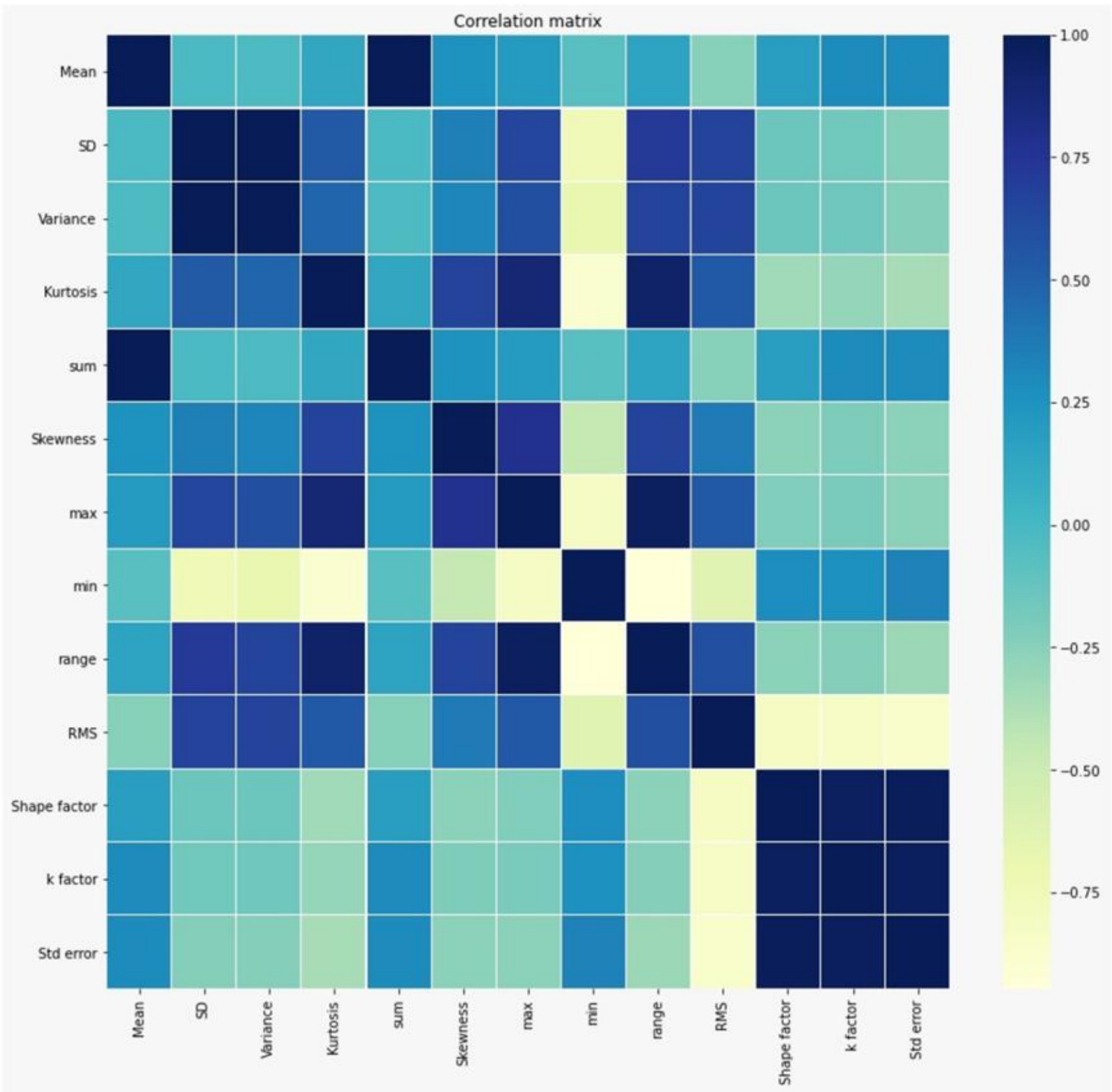

**Figure 4:** Correlation matrix between the top 13 features extracted using RFECV

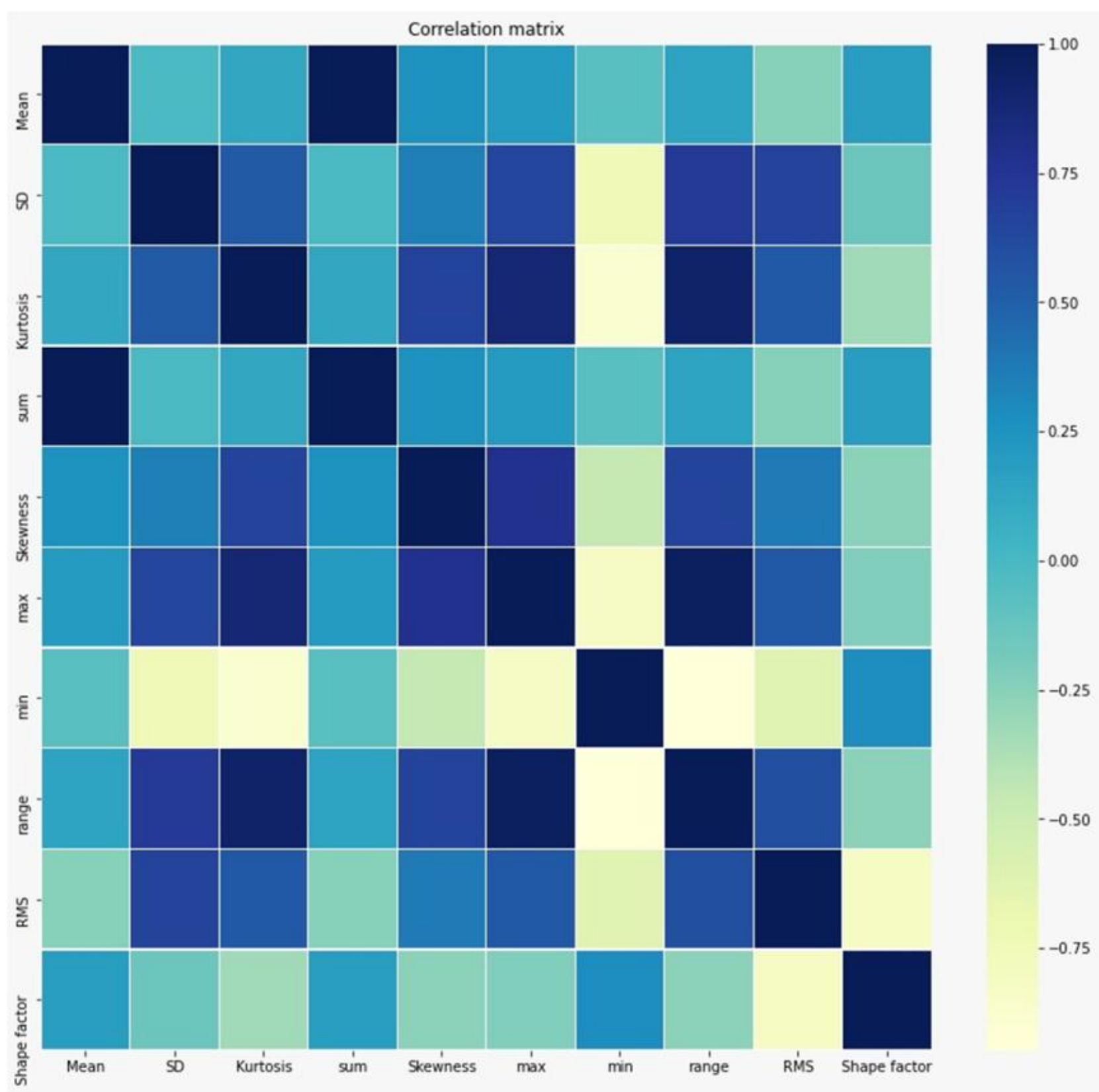

**Figure 5:** Correlation matrix between top 10 features (final features)\

Figure 6 shows the t-SNE plot of the final data. Figure 3 and Figure 6 show that the pre-processed data is much more superior and useful than the original data.

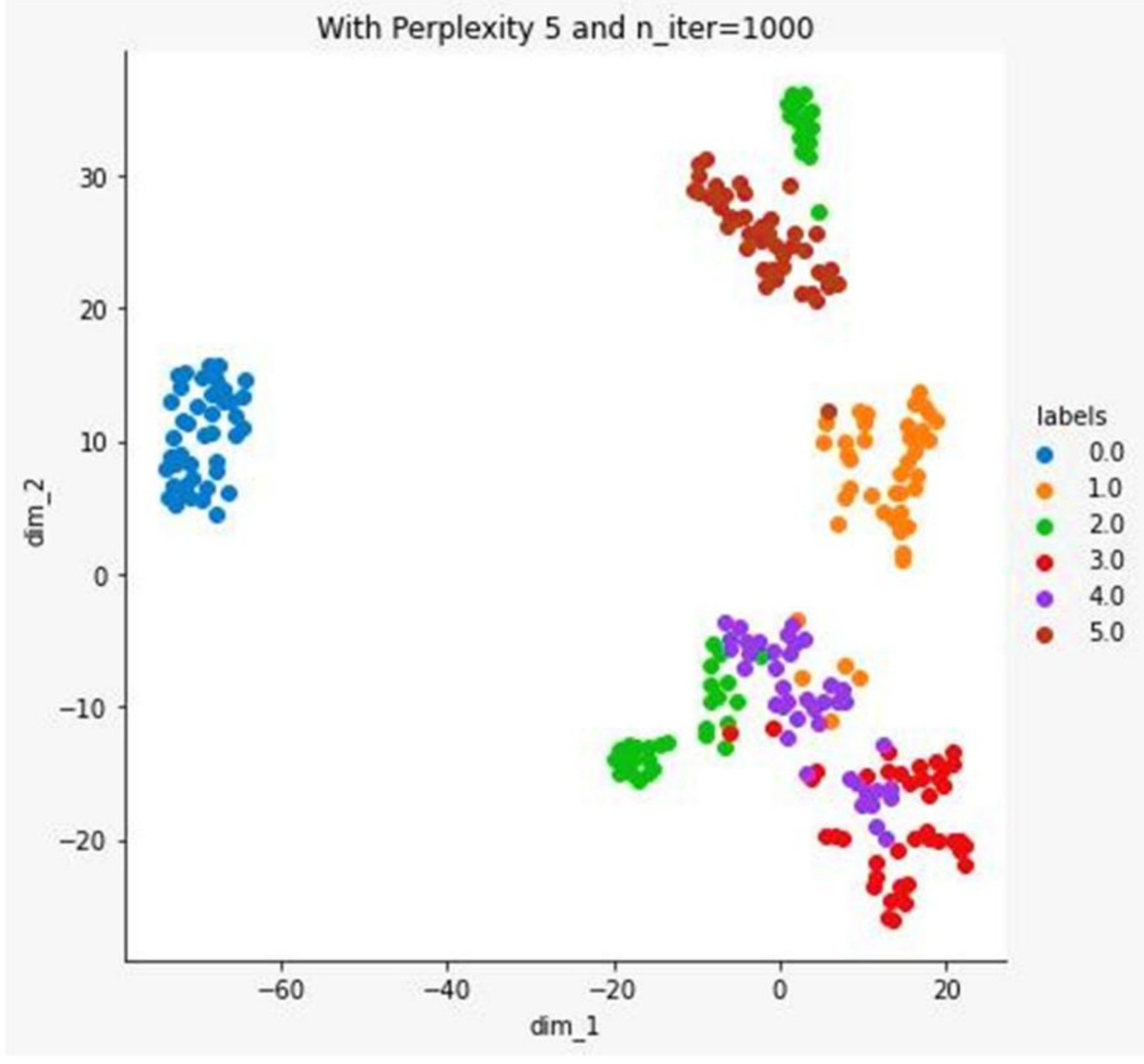

**Figure 6:** t-SNE plot of the final data (after standardization)

### 6.2 Classification using vanilla SVM

In this investigation, 250 samples were used for training and 50 for testing. Different SVM kernels like Linear, Sigmoid, Polynomial, and RBF were tested.

| | | | | | | | |
|---|---|---|---|---|---|---|---|
| Original tool health | NRML | 42 | 0 | 0 | 0 | 0 | 0 |
| | WNR | 0 | 37 | 0 | 0 | 3 | 2 |
| | WNT | 0 | 0 | 40 | 0 | 2 | 0 |
| | WFF | 0 | 0 | 1 | 38 | 2 | 0 |
| | WCRT | 0 | 0 | 1 | 3 | 37 | 0 |
| | WEGF | 0 | 1 | 2 | 0 | 0 | 39 |
| | | NRML | WNR | WNT | WFF | WCRT | WEGF |
| | | **Predicted tool health** | | | | | |

**Table 2:** Confusion matrix using vanilla SVM considering RBF kernel (10-fold CV)

| | | | | | | | |
|---|---|---|---|---|---|---|---|
| Original tool health | NRML | 8 | 0 | 0 | 0 | 0 | 0 |
| | WNR | 0 | 8 | 0 | 0 | 0 | 0 |
| | WNT | 0 | 0 | 7 | 0 | 1 | 0 |
| | WFF | 0 | 0 | 0 | 7 | 2 | 0 |
| | WCRT | 0 | 1 | 0 | 1 | 7 | 0 |
| | WEGF | 0 | 0 | 0 | 0 | 0 | 8 |
| | | NRML | WNR | WNT | WFF | WCRT | WEGF |
| | | **Predicted tool health** | | | | | |

**Table 3:** Confusion matrix using vanilla SVM considering RBF kernel (Testing)

Out of which RBF kernel (10-fold CV accuracy = 93.2%, Testing Accuracy = 90%) exhibited superior results with the default parameters. RBF kernel has 2 hyper-parameters for tuning. An RBF kernel is advantageous because of its similarity with the K-Nearest Neighbours algorithm (KNN). Further tuning of RBF kernel gives good results and so, it is used for further optimization of SVM's. Table 2 shows confusion matrix using vanilla SVM considering RBF kernel (10-fold CV). Table 3 shows confusion matrix using vanilla SVM considering RBF kernel (Testing). The abbreviations used for tool health are: Normal Condition: NRML, Wear at Nose Radius: WNR, Notch Wear: WNT, Wear at Flank Face: WFF, Crater Wear: WCRT, Edge Fracture: WEGF. Table 2 exhibits fully correct classification as there is no misclassification between healthy (NRML) and faulty tool. For, WNR tool condition 37 samples are correctly classified, 3 are misclassified as WCRT, and 2 are misclassified as WEGF. Similarly, classification and misclassification for other classes can be understood.

### 6.2.1 Design of Swarm-based Meta-heuristic Algorithms for TCM

In this study, SMO, Grid Search CV, Random Search CV, and Swarm-based meta-heuristic algorithms have been used for optimization of SVM. The results obtained are summarised in the following sections.

#### *6.2.1.1 Elephant Herding Optimization*

EHO is inspired by the herding behaviour of elephants as stated in. Elephants are social animals and the elephant groups are composed of several clans. The elephants belonging to different clans live together under the leadership of a matriarch, and male elephants remain solitary and will leave the family group while growing up. This habitation of elephants is used to solve optimization problems. Here the elephants are nothing but the SVM parameters C and $\gamma$. After initializing the size of the population (i.e., number of elephants and number of clans), the positions of the elephants are updated using 'clan updating operator' and 'clan separating operator'. A detailed designing and demonstration is presented for different applications [43-45].

- Clan Updating Operator

The elephant having the best fitness is the matriarch of that clan. All the elephants in the clan except the matriarch are updated at each generation using this formula:

$$X_{new,ci,j} = X_{ci,j} + \alpha \times (X_{best,ci} - X_{ci,j}) \times r \quad (1)$$

where $X_{new,ci,j}$ and $X_{ci,j}$ are the newly updated and old position for elephant j in clan $c_i$.

$\alpha \in [0,1]$ is a scale factor that determines the influence of matriarch ci on $X_{ci,j}$.

$r \in [0,1]$, is a uniform random distribution. $X_{best,ci}$ is the matriarch which is the fittest elephant individual in clan $c_i$.

The fittest elephant is updated by the following formula:

$$X_{new,ci,j} = \beta \times X_{center,ci} \quad (2)$$

where $\beta$ is a factor that determines the influence of $X_{center,ci}$ on $X_{new,ci,j}$ and,

$$X_{center,ci} = \frac{1}{n_{ci}} \times \sum_{j=1}^{n_{ci}} X_{ci,j,d} \quad (3)$$

where $1 \leq d \leq D$ indicates the $d^{th}$ dimension, and $D$ is the total dimensions.

- Clan Separating Operator

In elephant group, male elephants will leave their family group and will live alone when they reach puberty. This is idealized as taking the elephants having the worst fitness value and implementing the separation operator at each generation as shown below:

$$X_{worst,ci} = X_{min} + (X_{max} - X_{min} + 1) \times rand \quad (4)$$

where $X_{max}$ *and* $X_{min}$ are respectively upper and lower bound of the position of the elephant individual i.e., from the initialized parameters. $rand \in [0, 1]$, is a uniform random distribution. The results obtained after carrying out this procedure are tabulated and compared with other algorithms in Table 4.

*6.2.1.2 Monarch Butterfly Optimization*

MBO is inspired by the migration behavior of the Monarch Butterflies. Monarch butterflies stay in Land 1 (Northern USA) from April to August (5 months) and in Land 2 (Mexico) from September to March (7 months). Here the butterflies initialized are nothing but SVM parameters C and $\gamma$. After initializing the butterflies, they are separated into 2 subpopulations (NP) i.e. subpopulation-1 (NP-1) and subpopulation-2 (NP-2). NP-1 consists of butterflies having the best fitness value and the remaining butterflies are in NP-2.

The newly generated monarch butterfly replaces the old one only if it has a better fitness value. This process ensures that the fitness value after every iteration or generation does not deteriorate. A detailed designing and demonstration is presented for different applications [46-48]. The migration behavior of the Monarch Butterflies is idealized into following rules:

- Migration Operator

The generation of off-spring (position updating) is done by the migration operator. This can be adjusted by the migration ratio also called as the adjusting ratio ($p = 5/12$). This off-spring generation i.e., the position updating of the Monarch butterflies as they travel is given by the formula below:

$$X_{i,k}^{t+1} = X_{r1,k}^{t} \qquad if\ r \leq p \tag{5}$$

$$X_{i,k}^{t+1} = X_{r2,k}^{t} \qquad else \tag{6}$$

where, $X_{i,k}^{t+1}$ indicates the $k^{th}$ dimension of $X_i$ at generation $t + 1$. Parameters $r_1$ and $r_2$ are integer index that are randomly selected from NP-1 and NP-2 respectively. The parameter $r = rand \times peri$ where $rand \in [0,1]$, is a uniform random distribution and $peri$ represents the migration period ($peri = 1.2$).

- Butterfly Adjusting Operator

Tuning the position of other butterflies is also done by the Butterfly adjusting operator. The usage of butterfly adjusting operator mainly considers three factors as stated in [Monarch butterfly optimization a comprehensive review]:

- Global optimum is approached by the influence of social model.
- It takes into account the cognitive effects of other individuals by moving to a random individual
- The search scope and the population diversity are expanded by the use of Levy Flight

The new individuals of NP-2 are generated as below:

$$X_{i,k}^{t+1} = X_{best,k}^{t} \qquad if\ rand \leq p \tag{7}$$

$$X_{i,k}^{t+1} = X_{r3,k}^{t} \qquad if\ rand \geq p \tag{8}$$

Under this Eq. 8, there is another condition as given below:

$$X_{i,k}^{t+1} = X_{i,k}^{t} + \alpha \times (dx_k - 0.5) \qquad if\ rand \geq BAR \tag{9}$$

where, $X_{best,k}^{t}$ is the $kth$ element of the global optimum in generation $t$. The parameter $r_3$ is an integer index randomly selected from NP-2. BAR indicates Butterfly adjusting rate (7/12). The weighting factor $\alpha$ and $dx$ is given as:

$$\alpha = S_{max} / t^2 \tag{10}$$

$$dx = Levy(X_i^t) \tag{11}$$

Where, $S_{max}$ is the max walk step i.e., 10 and $Levy(X_i^t)$ is Levy function expressed as follows

$$Levy(X_i^t) = 0.01 \times \frac{u \times \sigma}{|v|^{\frac{1}{\beta}}},\ \sigma = \left(\frac{\Gamma(1+\beta) \times \sin(\frac{\pi\beta}{2})}{\Gamma(\frac{1+\beta}{2}) \times \beta \times 2^{\frac{\beta-1}{2}}}\right) \tag{12}$$

Here $u$ and $v$ random values from (0, 1), $\beta = 1.5$ and $\Gamma()$ is the gamma function.

After each iteration (i.e., generation), 10-fold CV accuracy of the newly generated population or generation is evaluated. After the specified number of iterations, the best positions/parameters are used as optimized SVM parameters. The results obtained after carrying out this procedure are tabulated and compared with other algorithms in Table 4.

*6.2.1.3 Harris Hawks Optimization*

In this section the effectiveness of Harris Hawks Optimization (HHO) method in predicting milling tool failure is evaluated. The HHO method utilizes the strategies that Harris' hawks implement in capturing a Jack Rabbit. In this study, the hawks are the SVM parameters and the rabbit is the set of parameters giving the maximum 10-fold CV accuracy.

A detailed designing and demonstration is presented for different applications [49-51]. After initializing a given size of population of hawks, the positions are updated using suitable strategies. These strategies work in 2 phases – exploration and exploitation. The transition between these 2 phases is carried out using the escape energy of the rabbit. This escape energy is given as

$$E = 2E_0\left(1 - \frac{t}{T}\right) \tag{13}$$

$$\text{Where } E_0 = 2r_a - 1 \tag{14}$$

And the jump strength of the rabbit is defined as

$$J = 2(1 - r_b) \tag{15}$$

Here, $r_a$ and $r_b$ are random numbers between 0 and 1.

The exploration phase is carried out using the following equation if $|E| \geq 1$

$$X(t+1) = \begin{cases} X_{rand}(t) - r_1|X_{rand}(t) - 2r_2X(t)| & q \geq 0.5 \\ (X_{rabbit}(t) - X_m(t)) - r_3(LB + r_4(UB - LB)) & q < 0.5 \end{cases} \tag{16}$$

Where $X(t+1)$ is the updated position and $X(t)$ is the current position of a hawk, defined by two values ($C$ and $\gamma$), $X_{rabbit}(t)$ is the current position of the rabbit, also known as the best position, $X_m(t)$ is the average position of the current hawk population, $r_1, r_2, r_3, r_4$ and $q$ are random numbers between 0 and 1, $LB$ and $UB$ are lower and upper bounds of SVM parameters and $X_{rand}(t)$ represents a randomly selected hawk from the current population.

The average position $X_m(t)$ is evaluated using the equation

$$X_m(t) = \frac{1}{N}\sum_{i=1}^{N} X_i(t) \tag{17}$$

If $|E| < 1$ the exploitation phase is carried out using hard and soft besiege processes. A few more conditions are taken into account for this purpose. When a random number $(r)$ between 0 and 1 and $|E|$ both are greater than or equal to 0.5 Soft besiege is applied using the following equation

$$X(t+1) = \Delta X(t) - E|JX_{rabbit}(t) - X(t)| \tag{18}$$

Where $\Delta X(t) = X_{rabbit}(t) - X(t)$ (19)

If $r \geq 0.5$ and $|E| < 0.5$ Hard besiege is implemented as follows

$$X(t+1) = X_{rabbit}(t) - E|\Delta X(t)| \tag{20}$$

In cases where $r < 0.5$, if $|E| \geq 0.5$ Soft besiege with progressive rapid dives is utilized as shown in equations (21), (22) and (24), else equations (21), (23) and (24) are used for Hard besiege with progressive rapid dives.

$$X(t+1) = \begin{cases} Y, & F(Y) < F(X(t)) \\ Z, & F(Z) < F(X(t)) \end{cases} \tag{21}$$

For soft besiege with progressive rapid dives

$$Y = X_{rabbit}(t) - E|JX_{rabbit}(t) - X(t)| \tag{22}$$

For hard besiege with progressive rapid dives

$$Y = X_{rabbit}(t) - E|JX_{rabbit}(t) - X_m(t)| \tag{23}$$

In both cases,

$$Z = Y + S \times LF(dim) \tag{24}$$

Where $S$ is any randomly selected number from (0, 1) and $LF(dim)$ is defined by equation (25)

$$LF(dim) = 0.01 \times \frac{u\times\sigma}{|v|^{\beta}}, \ \sigma = \left(\frac{\Gamma(1+\beta)\times\sin(\frac{\pi\beta}{2})}{\Gamma(\frac{1+\beta}{2})\times\beta\times 2^{\frac{\beta-1}{2}}}\right) \tag{25}$$

Here $u$ and $v$ random values from (0, 1), $\beta = 1.5$, $\Gamma()$ is the gamma function and $1 \times dim$ is the shape of $LF$. After each iteration the fitness i.e., 10-fold CV accuracy of the newly updated population is evaluated and the best position obtained at the end is used for the optimized SVM parameters.

In this study, population sizes $(N)$ of 50, 100, 150, 200 and 250 were evaluated for maximum 50 generations $(T)$ using $dim = 2$. The results obtained after carrying out this procedure are tabulated and compared with other algorithms in Table 4.

#### *6.2.1.4 Slime Mould Algorithm*

Slime Mould Algorithm utilizes the foraging technique of Physarum polycephalum. Foraging happens in 3 stages – approach and wrapping. The slime mould uses odour in approaching the food source. This behavior is expressed with the help of equation (26). Similar to HHO, SMA also expresses the location of slime mould in terms of SVM parameters. A detailed designing and demonstration is presented for different applications [52-54].

$$X(t+1) = \begin{cases} X_b(t) + vb(W \cdot X_A(t) - X_B(t)), & r_1 < p \\ vc \cdot X(t), & r_1 \geq p \end{cases} \tag{26}$$

Where $X_b(t)$ the current location of highest concentration of odour is while, $X(t)$ and $X(t + 1)$ is the current and updated location of the slime mould. $X_A(t)$ and $X_B(t)$ are randomly selected locations at each iteration, $vb$ is a randomly selected parameter lying in the interval $[-a, a]$ where $a$ is

$$a = \operatorname{arctanh}\left(-\left(\frac{t}{T}\right) + 1\right) \tag{27}$$

$vc$ is also a parameter that linearly decreases from 1 to 0 and $p$ is expressed as shown in equation (28)

$$p = \tanh|S(i) - DF|, \quad i = 1, 2, \ldots\ldots N \tag{28}$$

Where $S(i)$ is the fitness of $X(t)$ and $DF$ represents the best fitness calculated after all iterations.

W is the weight of slime mould expressed as

$$W(i) = \begin{cases} 1 + r_2 \log\left(\frac{bF - S(i)}{bF - wF} + 1\right), & i \leq N/2 \\ 1 - r_2 \log\left(\frac{bF - S(i)}{bF - wF} + 1\right), & \frac{N}{2} < i < N \end{cases} \tag{29}$$

$r_1$ and $r_2$ are random values from [0, 1]. Another parameter $z$ is initialized between [0, 0.1] to alter between approach and wrap stage. Taking into account the wrap stage, equation (26) is modified into the following equation where $r_3$ is also a random number from [0, 1].

$$X(t+1) = \begin{cases} r_3(UB - LB) + LB, & r_3 < z \\ X_b(t) + vb(W \cdot X_A(t) - X_B(t)), & r_1 < p \\ vc \cdot X(t), & r_1 \geq p \end{cases} \tag{30}$$

The Slime mould migrates in search of food using a venous network. Cytoplasm flows through this venous network giving a positive-negative feedback about the food sources. The flow of cytoplasm through a vein increases as it approaches a food source and the vein becomes thicker. This positive-negative feedback is achieved through propagation wave that is produced by a biological oscillator.

The weight W mathematically models the frequency of this oscillation. The parameters $vb$ and $vc$ model the selective behaviour of slime mould, that either approaches the food sources or searches for other sources. The results obtained after carrying out this procedure are tabulated and compared with other algorithms in Table 4.

*6.2.1.5 Moth Search Algorithm*

Researchers have studied and tried to understand scattered and self-organized systems that exhibit a collective behavior in nature. Numerous problems like planning, navigation, big data optimization are dealt with using this approach. The moth search algorithm is a general-purpose meta-heuristic approach. The flight patterns of moths are a representation of an optimization problem.

The moths having smaller distances from the best one always fly around their positions in Levy flight pattern. On the other hand, the best moth is at a distance from and is the target for all moths. Their flight towards the best is in line on account of photo taxis. This captures the two-fold nature of the algorithm viz. exploration and exploitation phase i.e. local and global search respectively.

The optimization problem of optimization algorithms is an intricate problem of balancing the exploitation and exploration phases. MSA has been benchmarked by twenty-five standard test functions and seven real-world applications in comparison with five other state-of-the-art meta-heuristic algorithms to demonstrate its effectiveness and performance. A detailed designing and demonstration is presented for different applications [55-57].

- Levy Flights

Lévy flight pattern is exhibited by moths that have a smaller distance from the best one. Mathematically,

$$y_i^{t+1} = y_i^t + \beta L(s) \tag{31}$$

$y_i^{t+1}$ - position of moth $i$ at time step $t+1$

$y_i^t$ - position of moth $i$ at time step $t$

$\beta$ - scale factor given as

$$\beta = \frac{S_{max}}{t^2} \tag{32}$$

$S_{max}$ is the max walk step and can be defined by the user based on the nature of the problem.

$L(s)$ - Lévy distribution

$$L(s) = \frac{(\sigma-1)\Gamma(\sigma-1)\sin\left(\frac{\pi(\sigma-1)}{2}\right)}{\pi s^{\sigma}} \tag{33}$$

Lévy flights of moths can be drawn from the Lévy distribution with $\sigma$ = 1.5[Wang]

- Fly straightly

The moths that are distant from the source of light fly towards it in a straight line. Mathematically,

$$y_i^{t+1} = \lambda \left(y_i^t + \varphi \left(y_{i\,best}^t - y_i^t\right)\right) \tag{34}$$

$y_{i\,best}^t$ - best moth at generation $t$

$\varphi$ - acceleration factor, set to golden ratio

$\lambda$ - scale factor which controls convergence speed, selected randomly from uniform standard distribution

But another possibility is moths may take a straight flight path to a destination beyond the light source. In that case, the final position of the moth is:

$$y_i^{t+1} = \lambda \left(y_i^t + \frac{1}{\varphi}\left(y_{i\,best}^t - y_i^t\right)\right) \tag{35}$$

We calculate updated positions with 50% possibility with equations (34) and (35). The results obtained after carrying out this procedure are tabulated and compared with other algorithms in Table 4.

### 6.4 Comparative Study

In table 4, the variable 'N' represents the number of instances initialized and 'T' means the number of epochs or generations, the optimizer was tested on. In most cases, the number of epochs is 50 because, after roughly 40 epochs, the optimizer stopped updating and exhibited more or less similar parameters.

Table 4 shows the comparison between results obtained using all swarm-based algorithms.

**Table 4:** Comparison between results obtained using all swarm-based algorithms

| Sr. No. | Metaheuristic algorithm | Common parameters | Algorithm specific parameters | C | γ | 10-fold CV accuracy | Training accuracy | Testing accuracy |
|---|---|---|---|---|---|---|---|---|
| 1. | EHO | $N = 30$<br>$T = 50$ | $clans = 6$<br>$\alpha = 0.5, \beta = 0.5$ | 2.60187031 | 0.19514524 | 95.2 | 97.6 | 90 |
| 2. | EHO | $N = 42$<br>$T = 50$ | $clans = 6$<br>$\alpha = 0.5, \beta = 0.5$ | 2.44192814 | 0.22250764 | 95.2 | 97.6 | 90 |
| 3. | EHO | $N = 42$<br>$T = 50$ | $clans = 6$<br>$\alpha = 0.9, \beta = 0.5$ | 2.12302576 | 0.2468275 | 95.2 | 97.6 | 86 |
| 4. | EHO | $N = 50$<br>$T = 50$ | $clans = 5$<br>$\alpha = 0.99, \beta = 0.5$ | 3.43610505 | 0.15436004 | 95.2 | 97.2 | 90 |
| 5. | EHO | $N = 50$<br>$T = 50$ | $clans = 5$<br>$\alpha = 0.99, \beta = 0.15$ | 2.21392735 | 0.24930654 | 95.2 | 97.6 | 86 |
| 6. | MBO | $N = 16$<br>$T = 50$ | $p = \frac{5}{12}, peri = 1.2$<br>$BAR = \frac{7}{12}, S_{max} = 10$ | 7.15540772 | 0.06674845 | 95.2 | 97.2 | 90 |
| 7. | MBO | $N = 16$<br>$T = 50$ | $p = \frac{5}{12}, peri = 1.2$<br>$BAR = \frac{7}{12}, S_{max} = 50$ | 12.15695916 | 0.05186617 | 95.2 | 97.2 | 90 |
| 8. | MBO | $N = 32$<br>$T = 50$ | $p = \frac{5}{12}, peri = 1.2$<br>$BAR = \frac{7}{12}, S_{max} = 50$ | 4.96573029 | 0.09577816 | 95.2 | 96.8 | 90 |

| Sr. No. | Metaheuristic algorithm | Common parameters | Algorithm specific parameters | C | γ | 10-fold CV accuracy | Training accuracy | Testing accuracy |
|---|---|---|---|---|---|---|---|---|
| 9. | MBO | $N = 32$<br>$T = 50$ | $p = \frac{4}{12}, peri = 1.2$<br>$BAR = \frac{7}{12}, S_{max} = 50$ | 1.540315659 | 0.25213978 | 94.8 | 96.8 | 86 |
| 10. | MBO | $N = 64$<br>$T = 50$ | $p = \frac{5}{12},$<br>$peri = 1.2$<br>$BAR = \frac{7}{12}, S_{max} = 10$ | 3.5599737 | 0.156438542 | 95.2 | 97.6 | 90 |
| 11. | HHO | $N = 50$<br>$T = 50$ | - | 6.25669968 | 0.06232359 | 95.6 | 97.2 | 90 |
| 12. | HHO | $N = 100$<br>$T = 50$ | - | 6.51794747 | 0.06375978 | 95.6 | 97.2 | 90 |
| 13. | HHO | $N = 150$<br>$T = 50$ | - | 6.30753555 | 0.06379345 | 95.6 | 97.2 | 90 |
| 14. | HHO | $N = 200$<br>$T = 50$ | - | 2.94113408 | 0.17820598 | 95.2 | 97.6 | 90 |
| 15. | HHO | $N = 250$<br>$T = 50$ | - | 6.69087997 | 0.06735923 | 95.6 | 97.2 | 90 |
| 16. | SMA | $N = 50$<br>$T = 50$ | $z = 0.03$ | 3.00311178 | 0.04710376 | 95.6 | 95.6 | 90 |
| 17. | SMA | $N = 50$<br>$T = 50$ | $z = 0.05$ | 4.81424431 | 0.03691801 | 95.6 | 96.4 | 90 |

| Sr. No. | Metaheuristic algorithm | Common parameters | Algorithm specific parameters | C | γ | 10-fold CV accuracy | Training accuracy | Testing accuracy |
|---|---|---|---|---|---|---|---|---|
| 18. | SMA | $N = 50$<br>$T = 50$ | $z = 0.1$ | 3.62436832 | 0.05771162 | 95.6 | 96.4 | 90 |
| 19. | SMA | $N = 100$<br>$T = 50$ | $z = 0.03$ | 6.35544493 | 0.03250747 | 95.6 | 96.4 | 90 |
| 20. | SMA | $N = 100$<br>$T = 50$ | $z = 0.1$ | 2.47431066 | 0.05532482 | 95.6 | 95.6 | 90 |
| 21. | MSA | $N = 50$<br>$T = 50$ | $S_{max} = 0.1$ | 300 | 0.00957350201 | 94.8 | 96.8 | 92 |
| 22. | MSA | $N = 50$<br>$T = 50$ | $S_{max} = 0.5$ | 300 | 0.00507901663 | 94.8 | 96.8 | 90 |
| 23. | MSA | $N = 50$<br>$T = 50$ | $S_{max} = 0.75$ | 300 | 0.00513813576 | 94.8 | 96.8 | 90 |
| 24. | MSA | $N = 100$<br>$T = 50$ | $S_{max} = 0.1$ | 4.99795924 | 0.100356 | 95.2 | 96.8 | 90 |
| 25. | MSA | $N = 100$<br>$T = 50$ | $S_{max} = 0.5$ | 300 | 0.00532959973 | 94.8 | 96.8 | 90 |
| 26. | Grid Search CV | 64 candidates | - | 1 | 1 | 92 | 98 | 84 |
| 27. | Random Search | 10 candidates | - | 10.64 | 1.063 | 91.19 | 100 | 84 |

### 6.5 White-box results

This section provides results for global and local interpretation of the highest and lowest performing model for both, optimized and vanilla SVM.

| Original tool health | | Predicted tool health | | | | | |
| --- | --- | --- | --- | --- | --- | --- | --- |
| | | NRML | WNR | WNT | WFF | WCRT | WEGF |
| | NRML | 42 | 0 | 0 | 0 | 0 | 0 |
| | WNR | 0 | 38 | 0 | 0 | 3 | 1 |
| | WNT | 0 | 0 | 41 | 0 | 1 | 0 |
| | WFF | 0 | 0 | 1 | 39 | 1 | 0 |
| | WCRT | 0 | 1 | 0 | 2 | 38 | 0 |
| | WEGF | 0 | 1 | 0 | 0 | 0 | 41 |

**Table 5:** Confusion matrix using Best Optimized - HHO algorithm driven SVM (10-fold CV)

| Original tool health | | Predicted tool health | | | | | |
| --- | --- | --- | --- | --- | --- | --- | --- |
| | | NRML | WNR | WNT | WFF | WCRT | WEGF |
| | NRML | 8 | 0 | 0 | 0 | 0 | 0 |
| | WNR | 0 | 7 | 0 | 0 | 1 | 0 |
| | WNT | 0 | 0 | 7 | 0 | 1 | 0 |
| | WFF | 0 | 0 | 0 | 8 | 1 | 0 |
| | WCRT | 0 | 1 | 0 | 1 | 7 | 0 |
| | WEGF | 0 | 0 | 0 | 0 | 0 | 8 |

**Table 6:** Confusion matrix using Best Optimized - HHO algorithm driven SVM (Testing)

The best optimized SVM has been achieved by using HHO where C = 6.51794747 and gamma = 0.06375978. The vanilla SVM is the one with default settings and kernel = ‘RBF’. Local representation is demonstrated by selecting a data point which has been correctly classified by the best optimized SVM and incorrectly classified by the optimized SVM. This has been done in order to compare the reasoning of two models while predicting a given data point. Table 5 shows confusion matrix using Best Optimized - HHO algorithm driven SVM (10-fold CV). Table 6 shows confusion matrix using Best Optimized - HHO algorithm driven SVM (Testing).

### 6.5.1 Global representation (Permutation Model) for best optimized SVM

| Weight | Feature |
|---|---|
| 0.4400 ± 0.0716 | RMS |
| 0.3720 ± 0.0742 | range |
| 0.2440 ± 0.1588 | sum |
| 0.1520 ± 0.0543 | Kurtosis |
| 0.1200 ± 0.0438 | min |
| 0.0160 ± 0.0299 | SD |
| 0 ± 0.0000 | Shape factor |
| 0 ± 0.0000 | max |
| 0 ± 0.0000 | Mean |
| -0.0080 ± 0.0408 | Skewness |

**Figure 7:** Permutation Model result for best optimized SVM

The permutation model will shuffle the value many times and will give the output average importance and its standard deviation as shown in Figure 7. While using the optimized model on the new data, to predict the condition of the tipped milling cutter, the most important thing the model must know is the RMS value and the range of the given data point. Figure 8 is the white-box decision tree model extracted from Best Optimized - HHO algorithm driven SVM.

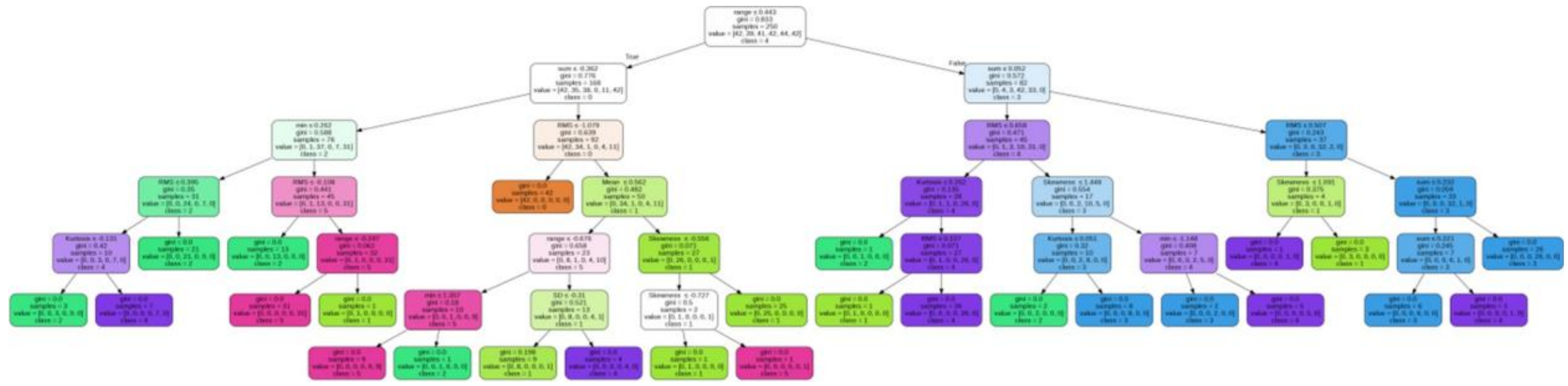

**Figure 8:** White-box decision tree model extracted from Best Optimized - HHO algorithm driven SVM

### 6.5.2 Global representation (Permutation Model) for vanilla SVM

| Weight | Feature |
|---|---|
| 0.4040 ± 0.1197 | range |
| 0.1840 ± 0.0588 | RMS |
| 0.1600 ± 0.0669 | Kurtosis |
| 0.1520 ± 0.1255 | sum |
| 0.1320 ± 0.0408 | SD |
| 0.1000 ± 0.0669 | Shape factor |
| 0.1000 ± 0.0912 | min |
| 0.0280 ± 0.0196 | max |
| 0 ± 0.0000 | Skewness |
| 0 ± 0.0000 | Mean |

**Figure 9:** Permutation Model result for vanilla SVM

Figure 9 shows that, the most important thing for the model to predict the condition of the tipped milling cutter is the range value. Figure 10 is the white-box decision tree model extracted from vanilla SVM.

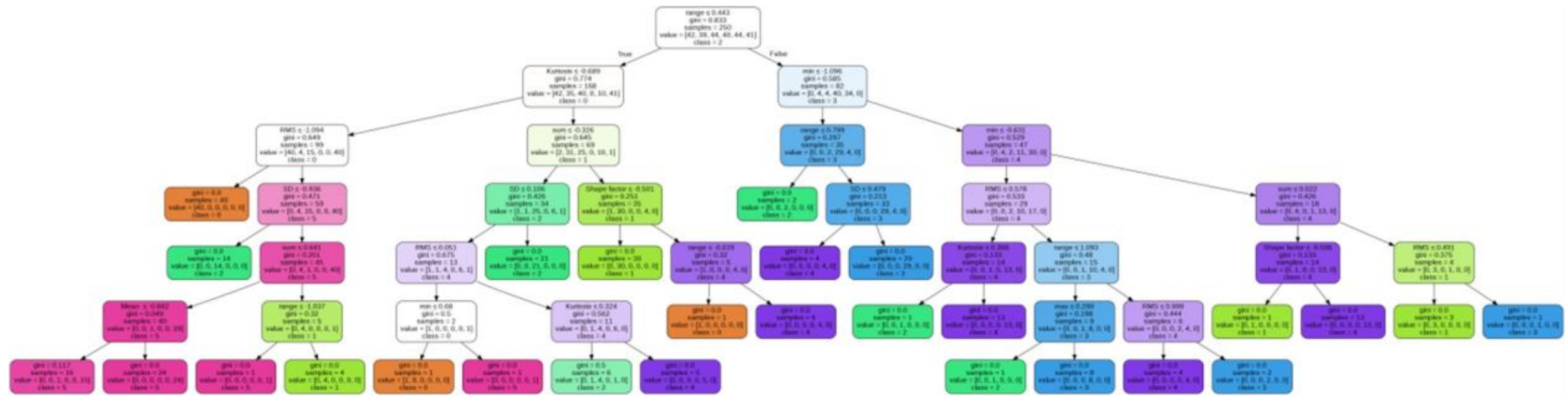

**Figure 10:** White-box decision tree model extracted from vanilla SVM

### 6.5.3 Local Representation for best optimized SVM

The true class label for this data point is 3 (Flank wear) and the best optimized SVM predicted it correctly. Table 7 helps to measure how important a feature is, not just based on the global score, but how each feature contributes to the model's decision-making process. The best optimized SVM is predicting class 3 with 100% probability. The Table 7 shows only the top features. The “sum”, “range”, “RMS”, and “<BIAS>” are the features contributing to the prediction result. Contribution column given in Table 7 above demonstrate the percentage of each feature contributed to the prediction result. The “<BIAS>” term is nothing but the expected average score output by the model. The feature value of “sum” is 0.391 and, having this particular value is forcing the model to predict the data point as class 3. The same value of “sum” is contributing a little for y=1 and has negligible contribution for y=0 and y=5 as it is not included in the top features.

| **y=0** (probability **0.000**) top features | | | **y=1** (probability **0.000**) top features | | |
|---|---|---|---|---|---|
| **Contribution** | **Feature** | **Value** | **Contribution** | **Feature** | **Value** |
| 0.168 | <BIAS> | 1 | 0.156 | <BIAS> | 1 |
| -0.168 | range | 0.758 | 0.032 | sum | 0.391 |
| | | | -0.081 | RMS | 0.516 |
| | | | -0.107 | range | 0.758 |
| **y=2** (probability **0.000**) top features | | | **y=3** (probability **1.000**) top features | | |
| **Contribution** | **Feature** | **Value** | **Contribution** | **Feature** | **Value** |
| 0.164 | <BIAS> | 1 | 0.383 | sum | 0.391 |
| -0.037 | sum | 0.391 | 0.344 | range | 0.758 |
| -0.127 | range | 0.758 | 0.168 | <BIAS> | 1 |
| | | | 0.105 | RMS | 0.516 |
| **y=4** (probability **0.000**) top features | | | **y=5** (probability **0.000**) top features | | |
| **Contribution** | **Feature** | **Value** | **Contribution** | **Feature** | **Value** |
| 0.226 | range | 0.758 | 0.168 | <BIAS> | 1 |
| 0.176 | <BIAS> | 1 | -0.168 | range | 0.758 |
| -0.024 | RMS | 0.516 | | | |
| -0.379 | sum | 0.391 | | | |

**Table 7:** Local representation of a single data point on best optimized SVM

### 6.5.4 Local representation for vanilla SVM

The true class label for this data point is 3 (Flank wear) and the vanilla SVM is incorrectly classifying it as class 4 (Crater Wear). Table 8 shows the top features only, which are governing the decision for the particular class. The vanilla SVM is predicting class 4 with 100% probability. “RMS’, “range”, “<BIAS>”, “min”, and “Kurtosis” are features contributing to the prediction result. “min” and “RMS” are features in opposite direction for class 3. By seeing both the results on the same data point, the “min” feature is playing a crucial role for prediction in the best optimized SVM and at the same time, it is preventing the vanilla SVM model to predict class 3.

| **y=0** (probability **0.000**) top features | | | **y=1** (probability **0.000**) top features | | |
|---|---|---|---|---|---|
| **Contribution** | **Feature** | **Value** | **Contribution?** | **Feature** | **Value** |
| 0.168 | <BIAS> | 1 | 0.156 | <BIAS> | 1 |
| -0.168 | range | 0.758 | -0.049 | min | -0.924 |
| | | | -0.107 | range | 0.758 |
| | | | | | |
| | | | | | |
| **y=2** (probability **0.000**) top features | | | **y=3** (probability **0.000**) top features | | |
| **Contribution?** | **Feature** | **Value** | **Contribution?** | **Feature** | **Value** |
| 0.176 | <BIAS> | 1 | 0.328 | range | 0.758 |
| 0.02 | min | -0.924 | 0.16 | <BIAS> | 1 |
| 0.002 | RMS | 0.516 | -0.143 | min | -0.924 |
| -0.071 | Kurtosis | 1.593 | -0.345 | RMS | 0.516 |
| -0.127 | range | 0.758 | | | |
| **y=4** (probability **1.000**) top features | | | **y=5** (probability **0.000**) top features | | |
| **Contribution?** | **Feature** | **Value** | **Contribution?** | **Feature** | **Value** |
| 0.342 | RMS | 0.516 | 0.164 | <BIAS> | 1 |
| 0.239 | range | 0.758 | -0.164 | range | 0.758 |
| 0.176 | <BIAS> | 1 | | | |
| 0.172 | min | -0.924 | | | |
| 0.071 | Kurtosis | 1.593 | | | |

**Table 8:** Local representation of a single data point on vanilla SVM

### 6.6 Conclusion

A real-time vibration signature and swarm-based optimized SVM based supervision of toothed milling cutter investigated on CNC milling trainer has been appropriately demonstrated in this research article. The outcomes achieved showcase the presented scheme to be explainable, capable & suitable of tool condition classification.

- The anomalous moments of vibration evolved due to in-process tool failures (i.e., flank and nose wear, crater and notch wear, edge fracture) have been investigated through time-domain response of acceleration. The response was observed as periodic & cyclic owing to distinctive failures that have properly been aided for apt tool condition classification.
- The Recursive Feature Elimination with Cross-Validation (RFECV) with decision trees as the estimator has aptly selected features yielding desirable classification.
- Design of SVM classifiers and its optimization using five swarm-based optimizers found to be competent and yields correct classification between healthy and faulty tool health. There was not misclassification observed between healthy and faulty class thus endorsing the adequate training & optimization of the classifier.
- The comparative study reveals that the best optimized SVM (classification accuracy 97.2%) has been achieved by Harris Hawks Optimization where C was 6.51794747 and gamma was 0.06375978.
- The use of model agnostic method for model interpretation and extracting the rules that govern the model's decision has been successfully presented. The Decision Tree has offered systematic rules from underlying SVM model using the Eli5 library. This white-box approach has positively provided insight into the performance of machine learning models in tool condition monitoring.
- The framework can be deployed on edge-old machine tools comprising few resources in terms of software & instruments.
- This supervised-SVM-based classification strategy shall assist optimized and efficient usage of milling cutter, evade idle time, ensure the reliability of a machining, and many more.

**References**

1. Patange, A. D., Jegadeeshwaran, R., & Dhobale, N. C. (2019, October). Milling cutter condition monitoring using machine learning approach. In *IOP Conference Series: Materials Science and Engineering* (Vol. 624, No. 1, p. 012030). IOP Publishing.
2. Kuntoğlu, M., & Sağlam, H. (2021). Investigation of signal behaviors for sensor fusion with tool condition monitoring system in turning. *Measurement*, *173*, 108582.
3. Kumar, D. P., Muralidharan, V., & Ravikumar, S. (2022). Histogram as features for fault detection of multi point cutting tool–A data driven approach. *Applied Acoustics*, *186*, 108456.

4. Shewale, M. S., Mulik, S. S., Deshmukh, S. P., Patange, A. D., Zambare, H. B., & Sundare, A. P. (2019). Novel machine health monitoring system. In *Proceedings of the 2nd International Conference on Data Engineering and Communication Technology* (pp. 461-468). Springer, Singapore.
5. Tambake, N. R., Deshmukh, B. B., & Patange, A. D. (2021, July). Data Driven Cutting Tool Fault Diagnosis System Using Machine Learning Approach: A Review. In *Journal of Physics: Conference Series* (Vol. 1969, No. 1, p. 012049). IOP Publishing.
6. Patil, S. S., Pardeshi, S. S., Patange, A. D., & Jegadeeshwaran, R. (2021, July). Deep Learning Algorithms for Tool Condition Monitoring in Milling: A Review. In *Journal of Physics: Conference Series* (Vol. 1969, No. 1, p. 012039). IOP Publishing.
7. Bajaj, N. S., Patange, A. D., Jegadeeshwaran, R., Kulkarni, K. A., Ghatpande, R. S., & Kapadnis, A. M. (2021). A Bayesian optimized discriminant analysis model for condition monitoring of face milling cutter using vibration datasets. *Journal of Nondestructive Evaluation, Diagnostics and Prognostics of Engineering Systems*, *5*(2), 021002.
8. Kuntoğlu, M., Salur, E., Gupta, M. K., Sarıkaya, M., & Pimenov, D. Y. (2021). A state-of-the-art review on sensors and signal processing systems in mechanical machining processes. *The International Journal of Advanced Manufacturing Technology*, *116*(9), 2711-2735.
9. Patange, A. D., & Jegadeeshwaran, R. (2021). A machine learning approach for vibration-based multipoint tool insert health prediction on vertical machining centre (VMC). *Measurement*, *173*, 108649.
10. Khade, H. S., Patange, A. D., Pardeshi, S. S., & Jegadeeshwaran, R. (2021). Design of bagged tree ensemble for carbide coated inserts fault diagnosis. *Materials Today: Proceedings, 46(2),* 1283-1289.
11. Patange, A. D., & Jegadeeshwaran, R. (2021). Review on tool condition classification in milling: A machine learning approach. *Materials Today: Proceedings, 46(2),* 1106-1115
12. Mulik Sharad, S., Deshmukh Suhas, P., & Patange Abhishek, D. (2017). Review of vibration monitoring techniques using low cost sensors and microcontrollers. *J. Mechatron. Autom*, *4*(2).
13. Nalavade, S. P., Patange, A. D., Prabhune, C. L., Mulik, S. S., & Shewale, M. S. (2019). Development of 12 Channel Temperature Acquisition System for Heat Exchanger Using MAX6675 and Arduino Interface. In *Innovative Design, Analysis and Development*

*Practices in Aerospace and Automotive Engineering (I-DAD 2018)* (pp. 119-125). Springer, Singapore.

14. Patange, A. D., & Jegadeeshwaran, R. (2020). Application of bayesian family classifiers for cutting tool inserts health monitoring on CNC milling. *International Journal of Prognostics and Health Management*, *11*(2), 1-9.
15. Mohanraj, T., Yerchuru, J., Krishnan, H., Aravind, R. N., & Yameni, R. (2021). Development of tool condition monitoring system in end milling process using wavelet features and Hoelder's exponent with machine learning algorithms. *Measurement*, *173*, 108671.
16. Li, W., Wang, G. G., & Alavi, A. H. (2020). Learning-based elephant herding optimization algorithm for solving numerical optimization problems. *Knowledge-Based Systems*, *195*, 105675.
17. Nayak, M., Das, S., Bhanja, U., & Senapati, M. R. (2020). Elephant herding optimization technique based neural network for cancer prediction. *Informatics in Medicine Unlocked*, *21*, 100445.
18. Li, J., Li, W., Chang, X., Sharma, K., & Yuan, Z. (2020). Real-time predictive control for chemical distribution in sewer networks using improved elephant herding optimization. *IEEE Transactions on Industrial Informatics*.
19. Dhillon, S. S., Agarwal, S., Wang, G. G., & Lather, J. S. (2020). Automatic generation control of interconnected power systems using elephant herding optimization. In *Intelligent Computing Techniques for Smart Energy Systems* (pp. 9-18). Springer, Singapore.
20. Li, W., & Wang, G. G. (2021). Improved elephant herding optimization using opposition-based learning and K-means clustering to solve numerical optimization problems. *Journal of Ambient Intelligence and Humanized Computing*, 1-32.
21. Yi, J. H., Wang, J., & Wang, G. G. (2019). Using Monarch butterfly optimization to solve the emergency vehicle routing problem with relief materials in sudden disasters. *Open Geosciences*, *11*(1), 391-413.
22. Bacanin, N., Bezdan, T., Tuba, E., Strumberger, I., & Tuba, M. (2020). Monarch butterfly optimization based convolutional neural network design. *Mathematics*, *8*(6), 936.
23. Singh, P., Meena, N. K., Yang, J., Vega-Fuentes, E., & Bishnoi, S. K. (2020). Multi-criteria decision making monarch butterfly optimization for optimal distributed energy resources mix in distribution networks. *Applied Energy*, *278*, 115723.

24. Ghanem, W. A., & Jantan, A. (2020). Training a neural network for cyberattack classification applications using hybridization of an artificial bee colony and monarch butterfly optimization. *Neural Processing Letters*, *51*(1), 905-946.
25. Dorgham, O. M., Alweshah, M., Ryalat, M. H., Alshaer, J., Khader, M., & Alkhalaileh, S. (2021). Monarch butterfly optimization algorithm for computed tomography image segmentation. *Multimedia Tools and Applications*, 1-34.
26. Houssein, E. H., Hosney, M. E., Oliva, D., Mohamed, W. M., & Hassaballah, M. (2020). A novel hybrid Harris hawks optimization and support vector machines for drug design and discovery. *Computers & Chemical Engineering*, *133*, 106656.
27. Elgamal, Z. M., Yasin, N. B. M., Tubishat, M., Alswaitti, M., & Mirjalili, S. (2020). An improved harris hawks optimization algorithm with simulated annealing for feature selection in the medical field. *IEEE Access*, *8*, 186638-186652.
28. Golilarz, N. A., Addeh, A., Gao, H., Ali, L., Roshandeh, A. M., Munir, H. M., & Khan, R. U. (2019). A new automatic method for control chart patterns recognition based on ConvNet and harris hawks meta heuristic optimization algorithm. *Ieee Access*, *7*, 149398-149405.
29. Bao, X., Jia, H., & Lang, C. (2019). A novel hybrid harris hawks optimization for color image multilevel thresholding segmentation. *Ieee Access*, *7*, 76529-76546.
30. Tikhamarine, Y., Souag-Gamane, D., Ahmed, A. N., Sammen, S. S., Kisi, O., Huang, Y. F., & El-Shafie, A. (2020). Rainfall-runoff modelling using improved machine learning methods: Harris hawks optimizer vs. particle swarm optimization. *Journal of Hydrology*, *589*, 125133.
31. Mostafa, M., Rezk, H., Aly, M., & Ahmed, E. M. (2020). A new strategy based on slime mould algorithm to extract the optimal model parameters of solar PV panel. *Sustainable Energy Technologies and Assessments*, *42*, 100849.
32. Zubaidi, S. L., Abdulkareem, I. H., Hashim, K. S., Al-Bugharbee, H., Ridha, H. M., Gharghan, S. K., ... & Al-Khaddar, R. (2020). Hybridised artificial neural network model with slime mould algorithm: a novel methodology for prediction of urban stochastic water demand. *Water*, *12*(10), 2692.
33. Kumar, C., Raj, T. D., Premkumar, M., & Raj, T. D. (2020). A new stochastic slime mould optimization algorithm for the estimation of solar photovoltaic cell parameters. *Optik*, *223*, 165277.
34. Liu, L., Zhao, D., Yu, F., Heidari, A. A., Ru, J., Chen, H., ... & Pan, Z. (2021). Performance optimization of differential evolution with slime mould algorithm for multilevel breast cancer image segmentation. *Computers in Biology and Medicine*, *138*, 104910.

35. Tiachacht, S., Khatir, S., Le Thanh, C., Rao, R. V., Mirjalili, S., & Wahab, M. A. (2021). Inverse problem for dynamic structural health monitoring based on slime mould algorithm. *Engineering with Computers*, 1-24.
36. Razmjooy, N., Razmjooy, S., Vahedi, Z., Estrela, V. V., & de Oliveira, G. G. (2021). A new design for robust control of power system stabilizer based on Moth search algorithm. In *Metaheuristics and Optimization in Computer and Electrical Engineering* (pp. 187-202). Springer, Cham.
37. Shankar, K., Perumal, E., & Vidhyavathi, R. M. (2020). Deep neural network with moth search optimization algorithm based detection and classification of diabetic retinopathy images. *SN Applied Sciences*, *2*(4), 1-10.
38. Singh, P., Bishnoi, S. K., & Meena, N. K. (2019). Moth search optimization for optimal DERs integration in conjunction to OLTC tap operations in distribution systems. *IEEE Systems Journal*, *14*(1), 880-888.
39. Carrasco, Ó., Crawford, B., Soto, R., Lemus-Romani, J., Astorga, G., & Salas-Fernández, A. (2019, June). Optimization of Bridges Reinforcements with Tied-Arch Using Moth Search Algorithm. In *International Work-Conference on the Interplay Between Natural and Artificial Computation* (pp. 244-253). Springer, Cham.
40. Han, X., Yue, L., Dong, Y., Xu, Q., Xie, G., & Xu, X. (2020). Efficient hybrid algorithm based on moth search and fireworks algorithm for solving numerical and constrained engineering optimization problems. *The Journal of Supercomputing*, *76*(12), 9404-9429.
41. Swagato Das and Purnachandra Saha (2021). Performance of swarm intelligence based chaotic meta-heuristic algorithms in civil structural health monitoring. Measurement. Vol. 169, 108533
42. S. Chunsheng, S. Erwei, H. Xiangyang, Z. Jinguang, Monitoring the cohesive damage of the adhesive layer in CFRP double-lapped bonding joint based on non-uniform strain profile reconstruction using dynamic particle swarm optimization algorithm, Measurement 123 (2018) 235–245.
43. Wang, G. G., Deb, S., & Coelho, L. D. S. (2015, December). Elephant herding optimization. In *2015 3rd International Symposium on Computational and Business Intelligence (ISCBI)* (pp. 1-5). IEEE.

44. Li, W., & Wang, G. G. (2021). Improved elephant herding optimization using opposition-based learning and K-means clustering to solve numerical optimization problems. *Journal of Ambient Intelligence and Humanized Computing*, 1-32.
45. Li, W., & Wang, G. G. (2021). Elephant herding optimization using dynamic topology and biogeography-based optimization based on learning for numerical optimization. *Engineering with Computers*, 1-29.
46. Wang, G. G., Deb, S., Zhao, X., & Cui, Z. (2018). A new monarch butterfly optimization with an improved crossover operator. *Operational Research*, *18*(3), 731-755.
47. Gai Ge Wang, Suash Deb and Zhihua Cui (2019). Monarch Butterfly Optimization. Neural Computing and Applications. vol. 31, 1995–2014.
48. Feng, Y., Deb, S., Wang, G. G., & Alavi, A. H. (2021). Monarch butterfly optimization: A comprehensive review. *Expert Systems with Applications*, *168*, 114418.
49. Song, S., Wang, P., Heidari, A. A., Wang, M., Zhao, X., Chen, H., ... & Xu, S. (2021). Dimension decided Harris hawks optimization with Gaussian mutation: Balance analysis and diversity patterns. *Knowledge-Based Systems*, *215*, 106425.
50. Shao, K., Fu, W., Tan, J., & Wang, K. (2021). Coordinated approach fusing time-shift multiscale dispersion entropy and vibrational Harris hawks optimization-based SVM for fault diagnosis of rolling bearing. *Measurement*, *173*, 108580.
51. Abdel-Basset, M., Ding, W., & El-Shahat, D. (2021). A hybrid Harris Hawks optimization algorithm with simulated annealing for feature selection. *Artificial Intelligence Review*, *54*(1), 593-637.
52. Li, S., Chen, H., Wang, M., Heidari, A. A., & Mirjalili, S. (2020). Slime mould algorithm: A new method for stochastic optimization. *Future Generation Computer Systems*, *111*, 300-323.
53. Abdel-Basset, M., Chang, V., & Mohamed, R. (2020). HSMA_WOA: A hybrid novel Slime mould algorithm with whale optimization algorithm for tackling the image segmentation problem of chest X-ray images. *Applied soft computing*, *95*, 106642.
54. Zubaidi, S. L., Abdulkareem, I. H., Hashim, K. S., Al-Bugharbee, H., Ridha, H. M., Gharghan, S. K., ... & Al-Khaddar, R. (2020). Hybridised artificial neural network model with slime mould algorithm: a novel methodology for prediction of urban stochastic water demand. *Water*, *12*(10), 2692.
55. Wang, G. G. (2018). Moth search algorithm: a bio-inspired metaheuristic algorithm for global optimization problems. *Memetic Computing*, *10*(2), 151-164.

56. Gai Ge Wang (2016). Moth search algorithm: a bio-inspired metaheuristic algorithm for global optimization problems. Memetic Computing. vol. 10, 151–164.
57. Feng, Y., & Wang, G. G. (2022). A binary moth search algorithm based on self-learning for multidimensional knapsack problems. *Future Generation Computer Systems*, *126*, 48-64.